\documentclass[letterpaper]{article} 
\usepackage{aaai24}  
\usepackage{times}  
\usepackage{helvet}  
\usepackage{courier}  
\usepackage[hyphens]{url}  
\usepackage{graphicx} 
\usepackage{natbib}  
\usepackage{caption} 
\usepackage{booktabs}
\usepackage{subcaption}
\usepackage{subcaption}
\usepackage{amsmath}
\usepackage{amsfonts}
\usepackage{enumitem}

\usepackage{algorithm}
\usepackage{algpseudocode}

\DeclareMathOperator*{\argmin}{argmin}
\DeclareMathOperator*{\argmax}{argmax}

\usepackage{mathtools} 

\usepackage{array}
\usepackage{makecell}
\usepackage{tablefootnote}

\newcommand{\algname}{\textsc{TCE-Search}}

\usepackage{newfloat}
\usepackage{listings}
\DeclareCaptionStyle{ruled}{labelfont=normalfont,labelsep=colon,strut=off} 
\floatstyle{ruled}
\newfloat{listing}{tb}{lst}{}
\floatname{listing}{Listing}
\title{A General Search-based Framework for Generating Textual Counterfactual Explanations}
\author {
    Daniel Gilo,
    Shaul Markovitch
}
\affiliations {
    Department of Computer Science \\
    Technion - Israel Institute of Technology \\
    $\{$danielgilo,shaulm$\}$@cs.technion.ac.il
}

\begin{document}

\frenchspacing
\maketitle

\begin{abstract}
One of the prominent methods for explaining 
the decision of a machine-learning classifier is by a counterfactual example.
Most current algorithms for generating such examples in the textual domain are
based on generative language models.  
Generative models, however, are trained to minimize a specific loss function in order to fulfill certain requirements for the generated texts.  Any change in the requirements may necessitate costly retraining, thus potentially limiting their applicability.
In this paper, we present a general search-based framework for generating counterfactual explanations in the textual domain.  
Our framework is model-agnostic,  domain-agnostic,  
anytime, and does not require retraining in order to adapt to changes in the user requirements. 
We model the task as a search problem in a space where the initial state is the classified text, and the goal state is a text in a given target class.  
Our framework includes domain-independent modification operators, but can also exploit domain-specific knowledge through specialized operators.
The search algorithm attempts to find a text from the target class with minimal user-specified distance from the original classified object. 
\end{abstract}

\section{Introduction}

An important feature of automatic decision making systems (DMS) is the ability to provide human-understandable explanations for their decisions.  Such explanations can be helpful for ensuring bias-free operation, for debugging the DMS, and for avoiding legal issues \citep{DBLP:conf/aaai/MadaanPPS21,doshi2017accountability}. 

Many DMS are based on models generated by machine-learning algorithms.  Some of the algorithms are based on \emph{interpretable} models, such as decision trees.
Others, 
however, are based on \emph{opaque} models, such as deep neural networks or randoms forests,
that, due to their complexity, are practically being used as a \emph{black box}, making explanation generation a challenging task.  

From research done in the social sciences \citep{miller2019explanation}, it is apparent that 
human-generated explanations are, often, contrastive. People do not explain why an event occurred per se, but rather explain why it had occurred compared to some other event which did not occur. Such explanations are called \emph{counterfactual explanations}. 

\citet{wachter2017counterfactual} presented the problem of generating counterfactual explanations for black-box classifications as an optimization task, where a counterfactual is an object that is classified differently from the original object by the black-box classifier, but is similar to it according to a chosen distance metric. Various algorithms have been proposed to address different forms of this optimization problem
(e.g. \citealp{karimi2020model, mothilal2020explaining,
galhotra2021explaining}).
The vast majority of them generate the counterfactuals by perturbing the input example in the feature space, 
yielding a feature vector.  Such an approach is not applicable, however,
in the textual domain, where objects are usually represented as vectors in some latent embedding spaces which are not comprehensible by humans.

To overcome these difficulties, several works have introduced an alternative approach that is based on generative models, yielding textual counterfactuals that are understandable by humans (e.g. \citealp{
yang2021generative,
DBLP:conf/acl/WuRHW20,
ross2020explaining, DBLP:conf/aaai/MadaanPPS21}).
This approach, however, has a fundamental limitation. Generative models are trained to minimize a specific loss function that is tailored to fulfill certain requirements for the generated texts. The challenge arises when these requirements need to be altered or modified across different use cases. For instance, there might be a shift from the need for visually similar counterfactuals to semantically similar ones, or from a requirement for fast response times to situations where more time is available. To accommodate such changes, generative methods often necessitate costly retraining processes to minimize a different loss function whenever the settings are modified.

In this paper, we present \algname\footnote{Textual Counterfactual Explanation using Search}, a general search-based framework for generating counterfactual explanation for text classification.
Our new method  can adapt to a range of user-specified constraints and preferences with no retraining required. These may include the distance function to be minimized, the required confidence threshold for counterfactual classification, and time constraints on the method's operation.

  \begin{table}
  \small
    \caption{Examples of \algname\ counterfactuals for a sentiment classification task.}
    \begin{tabular}{p{1.5in}p{1.5in}}
        \toprule
          Original Text                & Counterfactual      \\
        \midrule
          this beautifully animated epic is never dull. & this beautifully animated epic is never \textbf{entertaining}.         \\
          \midrule
          best punjabi food i've had in the north american continent            & \textbf{worst} punjabi food i've had in the north american continent         \\

        \bottomrule
    \end{tabular}
  \label{tab:example-cfs}
  \end{table}

Our method is based on modeling the counterfactual generation task as  a search problem.  The initial state is the classified text whose classification needs to be explained, and the goal state is a counterfactual example. 
We use a fine-tuned masked language model as well as a novel word replacement operator to replace parts of the text and explore the object space. Note that the fine-tuning process is performed only once, and changing the
user requirements does not necessitate to fine-tune again.
We also allow the use of text modification operators that are specific for 
particular problem types. The user requirements from the generated text
are encapsulated into a given distance function. Our algorithm searches for text from the user-defined target class with minimal distance from the original classified object. To accelerate the search, we introduce a heuristic function that is based on the confidence of the classification model.
Table \ref{tab:example-cfs} presents two simple examples for counterfactuals generated by \algname\ for sentiment classification tasks. Additional examples can be found in the appendix.

There are five important features that characterize our approach:

\begin{enumerate}[leftmargin=25pt]

    \item It is \emph{model-agnostic}.  While many existing approaches assume a specific learning model, such as neural networks,  our algorithm is independent of the type of model used.
    \item It is \emph{domain-independent} and can work for any text classification task using general modification operators, but also allows to exploit domain-specific knowledge, through specialized operators, for improved performance. 
    \item It can accommodate dynamic requirements on the generated text, such as the similarity criterion to the original text, without retraining.

    \item It can accommodate dynamic requirements on the generation process without retraining. 
    Specifically, it is an \emph{anytime} algorithm \citep{Zilberstein_1996},  meaning it can operate under various time constraints and provide improved solutions as more time is allocated.


    \item It can be applied to texts of any length.

\end{enumerate}

We demonstrate the effectiveness of our algorithm on various types of text classification problems with different user requirements, and show that it can generate counterfactual examples that are valid, close to the original example, and plausible.
We empirically show the advantage of our algorithm over state-of-the-art alternatives. Additionally, we report results of a human survey that validates the plausibility of our generated counterfactuals.

\section{Problem Definition}

Let $S$ be the set of all texts in English.
 Let $D \subseteq S$ be a sub-domain (e.g. movie reviews).
Let $C$ be a finite set of labels.
Let $\sigma: D \times C \rightarrow [0,1]$ be a model that estimates the probability of an object in $D$ to belong to a class in $C$, 
and $\phi(x) = \argmax_{y \in C}\sigma(x,y)$ be the associated classifier.

Let $x \in D$ be an object whose classification,
$y=\phi(x)$,
we need to explain.  Let $\hat{y} \in C \setminus \{y\}$ be 
a \emph{target label}\footnote{In the case of binary classification we can omit the specification of $\hat{y}$.}.
Let $\tau \in (0,1]$ be a classification confidence threshold.
We define $\hat{x} \in D$ to be a \emph{counterfactual example} to $x$ with respect to
$\hat{y}$ and $\tau$ if  $\phi(\hat{x}) = \hat{y} \land \sigma (\hat{x}, \hat{y}) > \tau$.  
 We denote the set of all such examples as $CF_{\tau, \hat{y}}(x)$.
This definition follows two common requirements \cite{verma2020counterfactual} from counterfactual examples: $\hat{x} \in D$ implies plausibility\footnote{Also appears in the literature as "fluency" or "likelihood" \cite{ross2020explaining, ross-etal-2022-tailor}}, and  $\phi(\hat{x}) = \hat{y}$ implies validity. 

 Given a counterfactual example $\hat{x}$, we adapt \citeauthor{wachter2017counterfactual}'s definition \citep{wachter2017counterfactual} to the textual domain, and define a \emph{counterfactual explanation} as 
a text of the form: "If  \emph{x} had been changed to $\hat{x}$, the classification
would have changed from $y$ to $\hat{y}$".
 As a \emph{counterfactual explanation} is immediate from a \emph{counterfactual example}, we use the two terms interchangeably.

Humans find counterfactual examples that are minimally changed, compared to the original example, easier to utilize as explanations \cite{thagard1989explanatory, miller2019explanation, Alvarez2019, wachter2017counterfactual}.
Therefore, given
a user-defined distance function between two texts $d: S \times S \rightarrow \mathbb{R}$, in addition to $\sigma$, $x$, $\hat{y}$ and $\tau$,
we can formulate our goal as an
optimization problem of finding $\argmin_{\hat{x} \in CF_{\tau, \hat{y}}(x)}d(x, \hat{x})$.

\section{Search-based Counterfactual Generation}
In this section, we present our new framework for the counterfactual generation task.  We first formulate it as a search problem:
\begin{enumerate}
    \item The set of states is $D$.  The initial state is $x$, an object whose classification we need to explain. 
    \item The goal predicate for $t \in D$ is $goal(t) \iff t \in CF_{\tau, \hat{y}}(x)$.  
    \item The set of operators is defined in the next subsections.
    \item The cost function $g$ is defined over states. Given a user-specified distance function $d$, and a state $t \in D$, $g(t) = d(x,t)$.  
\end{enumerate}

\subsection{Modification Operators}
\label{section:modification_operators}
To perturb the example in the object space, we utilize two types of modification operators: \emph{domain-agnostic} and \emph{domain-specific} that take a text, $x_{in}$, as input and produce a modified text - a candidate.
The first domain-agnostic operator is \textbf{mask filling}. In pre-processing time, a masked language model (MLM)\footnote{We have used DistilBERT \cite{sanh2019distilbert}.
} is fine-tuned for each class,
using the examples of this class from the training set.  During search, for each word $w$ in $x_{in}$, we swap it with a mask token, and apply the MLM, fine-tuned on the target class, to get $R(w)$, a set of replacements with their associated score. The operator returns the highest-score suggestions: $\hat{R}(w) = \{ (w',\xi') \in R(w) | \xi' \geq \alpha \cdot \max_{(\bar{w},\bar{\xi})\in R(w)} \bar{\xi} \}$ where $\xi'$ is the MLM score of the suggested replacement\footnote{ $0\leq \alpha \leq 1$ is a parameter set to 0.5 in our experiments.}.
Similarly, in order to add words to $x_{in}$, we insert a mask token between every two words in the text, and ask the MLM for suggested filling.
In addition, we utilize  a word-removal operator, that modifies $x_{in}$ by simply omitting words from it.

To produce a wider variety of goal-directed modifications, we introduce a second, novel domain-agnostic operator:  \textbf{differentiating words banks}. This operator utilizes words that were identified, during pre-processing, as differentiating a specific class from others, and replaces words in $x_{in}$ with words that are specifically associated with the target class.

To preserve grammatical correctness, we replace words with differentiating words of the same part of speech (POS). 
For a given POS, we create a bank of words for each class, that differentiate that class from the others. 
Given a word $w$ that belongs to that POS (and appears in a text from the training set) and a class $c$, we conduct a multinomial test to determine if $w$ is significantly over-represented in texts from $c$ ($p < 0.05$). The bank of each class consists of the $k$  over-represented words with the lowest $p$ value\footnote{We used k = 10.}. 

We observed that depending on the application domain, some POS are more likely
to separate the classes.  For example, 
"adjective" may be a differentiating POS
for movie review sentiment analysis, 
while "noun" is such for topic classification.
To reduce the branching factor of the search procedure, we therefore  first identify which parts of speech are the relevant ones, and construct differentiating words banks for those POS only.
For every POS and every class, we collect the list of words of that POS that appear in texts in the training set that belong to the class (e.g. all the verbs that appear in a positive example).  
We then use a semantic embedder\footnote{We used GloVe \cite{Pennington14}
.} to transform every word in each classes' lists to a vector space, and we use MANOVA\footnote{Some of MANOVA's assumptions 
may be violated.} to determine if the clusters' means are significantly different. POS where MANOVA test returns a significantly small $p$ value ($p<0.05$) are the differentiating POS.
For a given word $w$ in $x_{in}$, we check if its POS is a differentiating one, in which case we swap $w$ with each of the words in the bank associated with that POS in the target class.

In some domains, we may be able to utilize prior knowledge and define \emph{domain-specific} operators
that direct the search process towards a goal state. 
Namely, some of the experiments described in this paper deal with sentiment analysis problems.  The domain-specific operators we introduce for this task are based on WordNet antonyms \cite{miller1995wordnet}. For a given word in $x_{in}$, we swap it with its antonyms.

\subsection{Plausibility Enforcement}\label{section:plausibility_enforcement}
There is a possibility that our operators will take us out of $D$, violating the \emph{plausibility} requirement. 
To ensure plausibility, we use a fine-tuned language model (LM) 
\footnote{We used a fine-tuned GPT-2 \cite{radford2019language}
.} to filter out the candidates (produced by the modification operators) that are estimated to be out of $D$.
Since the original example $x$ is in $D$, we demand that every node in the search procedure is nearly as likely to appear in $D$ as $x$, i.e. nearly as \emph{plausible} as $x$. 
To this end, following \cite{ross2020explaining, ross-etal-2022-tailor}, we compute the language-modeling loss for $x$ and for each of the candidates, and return only those with a ratio of losses (candidate loss / original loss) that is at most $\gamma$, where $\gamma$ is a positive parameter \footnote{We used $\gamma = 1.5$ in our experiments.}. 
 Lower $\gamma$ values introduce a more strict plausibility threshold.

\subsection{The Heuristic Function}
Our heuristic function $h$ estimates the distance
of $t\in D$
to $CF_{\tau, \hat{y}}(x)$ by  
the normalized distance between the estimated target class probability for $t$, and the required classification confidence threshold.
Formally, $h(t)= \max \{0,\frac{\tau - \sigma(t, \hat{y})}{\tau}\}$.

\begin{algorithm}[t]
\caption{\algname}
\label{alg:outside_loop}
\hspace*{2pt}\textbf{Input:} $x, d, \hat{y}, \tau, TS$\hfill{$/*$ $TS$ is the training set$*/$}
\begin{algorithmic} 
\State $cf \gets \argmin_{\hat{x} \in TS \cap CF_{\tau, \hat{y}}(x)} d(x, \hat{x})$

\State $w_h \gets 1.0$

\While{ not interrupted}
\State $new\_cf \gets \mbox{WA}^*(w_h, d, x, \hat{y}, \tau)$
\If{$d(x, new\_cf) < d(x, cf)$}
    \State $cf \gets new\_cf$
\EndIf
\State $w_h \gets \frac{w_h}{2}$
\EndWhile
\State \Return $cf$
\end{algorithmic}
\end{algorithm}

\subsection{The Search Framework} \label{the_search_framework}
We base our framework on a variation of the weighted-$A^*$ algorithm \cite{pohl1970heuristic}, where the node selected to be expanded is the one with the lowest $f = (1-w_h)\cdot g + w_h\cdot h$ score where $w_h \in [0,1]$.
Contrary to the original weighted-A$^*$, we define the cost function over states rather than graph edges, and the heuristic function is not admissible. 
Our search framework works by iteratively calling our variation of weighted-A$^*$ (WA$^*$) with decreasing values of $w_h$. 
The algorithm maintains the best solution found so far and returns it when interrupted (by the user or by the clock). 
This makes our search an \emph{anytime} process. 

Algorithm \ref{alg:outside_loop} presents \algname. To ensure that a valid counterfactual is generated, the algorithm starts with a very fast baseline that
returns a counterfactual from the training set with the lowest $d$ (\emph{default counterfactual}).  If the training set is very large, we can return the closest counterfactual found in a sample of it. 
It then calls W$A^*$ with $w_h=1$,
and continues to call it, iteratively, while reducing $w_h$ in every iteration, reducing the effect of $h$.  Thus, each iteration is expected to generate a better solution than the previous at the expense of increased search time.
To save time, we cache the results of the calls to the MLM and LM and use it in the following iterations of the anytime algorithm.

\subsection{Handling Longer Texts}
As the branching factor of the search space is proportional to the length of the input text,
we introduce a focused version of the algorithm to deal with longer texts.
Let $t = \langle s_1, \ldots s_n\rangle$ be a longer
text where $s_i$ are sentences. 
The modified algorithm starts, as before, with $w_h=1$.
Instead of working on the entire text at once, 
the algorithm iteratively selects a sentence $s_i$ and
uses the basic algorithm to find a counterfactual $\hat{s_i}$ for
it.  If $\langle s_1, \ldots s_{i-1}, \hat{s_i}\ldots s_n\rangle \in CF_{\tau, \hat{y}}(t)$, or it runs out of sentences, the algorithm
proceeds to the next top-level iteration with a reduced $w_h$ as previously described. 

The sentences are selected according to their 
estimated importance evaluated by: 
  $\theta(s_i) = \sigma(t \setminus \{s_i\}, \hat{y}) - \sigma(t, \hat{y}) $,
where $t \setminus \{s_i\}$ is $t$ with the sentence $s_i$ removed. Intuitively, $\theta$ is an importance score for each sentence, measured by the amount of confidence gained in the target prediction once the sentence has been omitted from the input text.

\section{Empirical Evaluation}
We evaluated the performance of our framework on 
a variety of classification tasks.

\subsection{Experimental Methodology}
We report results for 8 datasets:
(1) \textbf{Yelp}. Yelp business reviews. (2) \textbf{Amazon} \citep{ni2019justifying}. Video game reviews on Amazon. (3) \textbf{SST} \citep{socher2013recursive}. Stanford sentiment treebank.
(4) \textbf{Science}. Comments from Reddit on scientific subjects. 
(5) \textbf{Genre}. Movie plot descriptions labeled by genre. 
(6) \textbf{AGNews} \citep{DBLP:conf/nips/ZhangZL15}. AG's news articles labeled by topic.
(7) \textbf{Airline}. Tweets about flight companies. (8) \textbf{Spam} \citep{DBLP:conf/doceng/AlmeidaHY11}. SMS labeled either as spam or as legitimate.
We applied a standard pre-processing procedure, the details of which can be found in the appendix.

For each dataset, we randomly sampled 200 examples  to serve as the explanation test set. We randomly split the rest of the dataset, so that 80\% serves as the black-box training set and 20\% remains for the black-box test set.

For each training set, we fine-tuned a LM to be used for plausibility enforcement, and a MLM for each class for the mask filling operator\footnote{We used the same fine-tuning procedure for both the LM and the MLM: 3 epochs with initial LR of 5e-05 and weight decay of 0.0 for AdamW. 
 We used a batch size of 2.}. Importantly, these models only require fine-tuning \textbf{once per training set}. Afterwards, we can utilize them in different scenarios, employing various distance functions that represent changing user requirements, without further training.

We consider 
 three distance criteria between the original text $x$ and the generated counterfactual $\hat{x}$: 
\begin{enumerate}
    \item Normalized word-level Levenshtein distance \cite{levenshtein1966binary} $\frac{d_{lev}(x, \hat{x})}{|x|}$, where $|x|$ is the number of words instances in $x$.
    \item Normalized cosine distance between $x$ and $\hat{x}$'s encodings\footnote{For encoding, we used the HuggingFace model available at: https://huggingface.co/sentence-transformers/all-MiniLM-L6-v2}: $\frac{d_{cosine}(enc(x), enc(\hat{x}))}{2}$.
    \item Normalized syntactic tree distance \cite{zhang1989simple} $\frac{d_{tree}(x, \hat{x})}{|x|}$. When dealing with texts longer than a single sentence, we sum the distances of the sentences. 
\end{enumerate}

As previously outlined, \algname\ is an anytime algorithm, which terminates its run when the allocated computational budget is exhausted.  A common method for measuring the resources consumed by a search algorithm is to count the number of nodes that the algorithm expanded during its execution.
 However, in our experiments we found that metric to be problematic, as the time it takes to expand a node varies significantly between nodes of different text length, as well as between cached (previously seen) nodes and uncached ones. 
 Since uncached calls to the LM and MLM took up more than 95\% of our algorithm's running time, we decided to use the number of such expensive calls (EC) as our metric for measuring the time resources consumed by the algorithm.

 In all the following experiments, unless otherwise stated, the following configuration was used as the standard setting:
 \begin{itemize}
     \item A random forest was used as the black box classifier. Training details and test sets performance can be found in the appendix.
     \item The full version of our framework was utilized, incorporating all operators introduced in section 3.1 to explore the search space.
     \item \algname\ was given a budget of 2000 expensive calls (EC).
     \item Levenshtein distance was used as the distance function, the results were averaged across the explanation test set.
     \item The classification confidence threshold $\tau$ is 0.5.
     \item No text length limitation was imposed on the datasets.
 \end{itemize}

 \begin{table*}
  \centering
  \small
  \caption{Average normalized Levenshtein (Lev), Cosine and Syntactic tree (Syn) distances, for sentiment analysis datasets with 256 words (256w) text length restriction. RoBERTa-LARGE model was used as the black box classifier. Best (lowest) results are bold. We use an asterisk (*) to indicate results that are statistically significantly different (paired t-test, $p=0.05$) from \algname.}
    \resizebox{0.9\textwidth}{!}{
    \begin{tabular}{lcccccccccccccc}
        \toprule

        \multicolumn{1}{c}{} & \multicolumn{3}{c}{{Yelp}$_{256w}$} & \multicolumn{3}{c}{Amazon$_{256w}$}
        & \multicolumn{3}{c}{Airline$_{256w}$}
        & \multicolumn{3}{c}{SST$_{256w}$}\\
        \cmidrule(rl){2-4}  \cmidrule(rl){5-7} \cmidrule(rl){8-10}  \cmidrule(rl){11-13}
        Method & Lev & Cosine & Syn & Lev & Cosine & Syn & Lev & Cosine & Syn & Lev & Cosine & Syn \\
         \toprule
         \algname\ & 0.56 &\textbf{0.066} &\textbf{0.121} & 0.448 &\textbf{0.087} &\textbf{0.144} &\textbf{0.257} &\textbf{0.047} &\textbf{0.145} &\textbf{0.207} &\textbf{0.043} &\textbf{0.222}
         \\
         MiCE &\textbf{0.266}* & 0.084* & 0.351* & \textbf{0.32}* & 0.106* & 0.386* & 0.269 & 0.1* & 0.289* & 0.256* & 0.11* & 0.336*
          \\
    \bottomrule
   \end{tabular}
   }
  \label{tab:vs_mice}
\end{table*}

\begin{table*}
  \centering
  \small
  \caption{Average normalized Levenshtein (Lev), Cosine and Syntactic tree (Syn) distances, for datasets with 1 sentence (1s) text length restriction. Best (lowest) results are bold. We use an asterisk (*) to indicate results that are statistically significantly different (paired t-test, $p=0.05$) from \algname.}
    \resizebox{0.8\textwidth}{!}{\begin{tabular}{lcccccccccccccc}
        \toprule

        \multicolumn{1}{c}{} & \multicolumn{3}{c}{{Yelp}$_{1s}$} & \multicolumn{3}{c}{Amazon$_{1s}$}
        & \multicolumn{3}{c}{Airline$_{1s}$}
       \\
        \cmidrule(rl){2-4}  \cmidrule(rl){5-7} \cmidrule(rl){8-10}  
        Method & Lev & Cosine & Syn & Lev & Cosine & Syn & Lev & Cosine & Syn \\
         \toprule
         \algname\ &\textbf{0.148} & \textbf{0.027} & \textbf{0.076} & \textbf{0.229} & \textbf{0.05} &
         \textbf{0.208} & \textbf{0.174} & \textbf{0.025} &
         \textbf{0.157} &
         \\
         Polyjuice & 0.692* & 0.16* & 0.542* & 0.591* & 0.163* & 0.596* & 0.723* & 0.168* & 0.657* 
          \\
          Tailor & 0.8* & 0.19* & 0.65* & 0.715* & 0.199* & 0.752* & 0.763* & 0.162* & 0.671* 
          \\
    \bottomrule
   \end{tabular}}
   
   \resizebox{0.8\textwidth}{!}{\begin{tabular} 
   {lcccccccccccccc}

        &\multicolumn{3}{c}{{SST}$_{1s}$} & \multicolumn{3}{c}{Science$_{1s}$}
        & \multicolumn{3}{c}{AGnews$_{1s}$}
       \\
        \cmidrule(rl){2-4}  \cmidrule(rl){5-7} \cmidrule(rl){8-10}  
         & Lev & Cosine & Syn& Lev & Cosine & Syn& Lev & Cosine & Syn\\
         \toprule
         \algname\ & \textbf{0.085} & \textbf{0.009} & \textbf{0.082} & \textbf{0.266} & \textbf{0.048} & \textbf{0.379} & \textbf{0.14} & \textbf{0.027} & \textbf{0.175} &
         \\
         Polyjuice & 0.562* & 0.159* & 0.653* & 0.678* & 0.199* & 0.818* & 0.799* & 0.313* & 0.891*
          \\
          Tailor & 0.658* & 0.174* & 0.784* & 0.716* & 0.218* & 0.903* & 0.818* & 0.311* & 0.948*
          \\
    \bottomrule
   \end{tabular}}

  \label{tab:vs_polyjuice}
\end{table*}

\subsection{Comparison to Existing Works}\label{section:comparison_with_existing_works}
We evaluate the performance of \algname\ compared to 3 baselines with published source code: \textbf{MiCE} \cite{ross2020explaining}, \textbf{Polyjuice} 
 \cite{DBLP:conf/acl/WuRHW20} and \textbf{Tailor} \cite{ross-etal-2022-tailor}. Further description of these works is in the Related Work section.
 
User requirements for generated counterfactuals can be formulated as a distance function from the original text. We compare the performance of the methods over 3 distance metrics, representing different possible user requirements.

All three baseline methods are based on generative models and are not guaranteed to
generate a valid counterfactual (i.e. a text
classified as the target class).  Additionally,
the models were trained to minimize a distance function that do not necessarily correspond to the one used to measure performance.  To address these issues, we use augmented versions of the 3 baseline models.  Each generative model is used to generate a set of 10 candidate texts.
The set is then filtered to include only valid counterfactuals,  and the one with the lowest distance (according to the distance function used for evaluation) is selected. Importantly, for our \algname\ algorithm, this procedure is not necessary, as it is guaranteed to generate a valid counterfactual and the distance function is used as the cost function for the search procedure, so it is directly optimized during the search.

To ensure fair comparison, all methods produce the \emph{default counterfactual} (see Section \ref{the_search_framework}) if they are unable to generate a closer valid counterfactual. 

Since MiCE is limited to differentiable models and requires access to the model's gradient, when comparing to MiCE, we used the exact model from the MiCE paper, RoBERTa-LARGE trained on the IMDB sentiment analysis dataset \cite{maas-EtAl:2011:ACL-HLT2011}\footnote{For full details on the model's architecture and training, please refer to MiCE paper.}, as the black box classifier. Since the model is trained on a sentiment analysis dataset, we use only the sentiment analysis datasets for comparison with MiCE. Furthermore, since MiCE is based on the T5 model \cite{raffel2020exploring} with a maximum input sequence length of 512 tokens, we 
used modified versions of the datasets for comparison 
with MiCE, omitting texts longer than 256 words, as a word may be composed of multiple tokens. Similarly, for comparison with Polyjuice and Tailor, which are designed to modify individual sentences, we created modified versions of the datasets containing only single sentences. We did not include the Genre and Spam datasets in this comparison becaues of the small number of single-sentence texts in them.

The comparisons results are available at Tables \ref{tab:vs_mice} and \ref{tab:vs_polyjuice}. 
The results show that \algname\ is able to produce counterfactuals that are generally closer to the original example than the alternatives, across different distance criteria, datasets, and text length limitations. This experiment displays \algname's ability to adapt to different user requirements, encapsulated in the distance function that was optimized in the search procedure, without retraining.  It is worth noting, however, that the baselines used fewer EC compared to \algname\footnote{To the best of our understanding, both Tailor and Polyjuice require a single forward pass (EC) to generate a modified text. As for MiCE, the authors state that the algorithm performs three "edit rounds", each costs approximately 360 EC, resulting in a total of around 1080 EC.}.

\subsection{The Effect of Different Modification Operators}

In order to evaluate the effectiveness of the operators introduced in Section \ref{section:modification_operators}, we compare 3 variations of our proposed framework: (1) \textbf{\algname-full} that uses all operators, (2) \textbf{\algname-no-DWB}, that does not use the differentiating words banks operator, and (3) \textbf{\algname-no-antonyms}, that uses only the domain-agnostic operators, without using the WordNet antonym operator.

Table \ref{tab:operators_effect} shows the performance of the algorithm variations on all eight datasets. The DWB operator had a statistically significant impact on performance, while the domain-specific antonyms operator's effect was not significant (paired t-test, $p=0.05$).

\begin{table*}[t]
  \centering
  \small
  \caption{Average normalized Levenshtein distance, for various \algname\ variations. Best (lowest) results are bold. We use an asterisk (*) to indicate results that are statistically significantly different ($p=0.05$) from the full version.}
    \resizebox{0.8\textwidth}{!}{\begin{tabular}{lccccccccc}
        \toprule

        Variation & SST & Yelp & Amazon & Genre & Spam & Airline & AGnews & Science\\
         \toprule
         \algname-full & 0.088 & \textbf{0.426} & \textbf{0.398} & \textbf{0.269} & 0.261 & \textbf{0.153} & \textbf{0.234} & \textbf{0.305}
         \\
         \algname-no-antonyms & \textbf{0.086} & 0.437 & 0.411 & 0.29 & \textbf{0.254} & 0.154 & 0.237 & 0.311
          \\
          \algname-no-DWB & 0.135* & 0.510* & 0.47* & 0.464* & 0.362* & 0.2* & 0.343* & 0.442*
          \\
    \bottomrule
   \end{tabular}}

  \label{tab:operators_effect}
\end{table*}

\begin{figure}[t]
    \centering
     \includegraphics[width=0.45\textwidth]{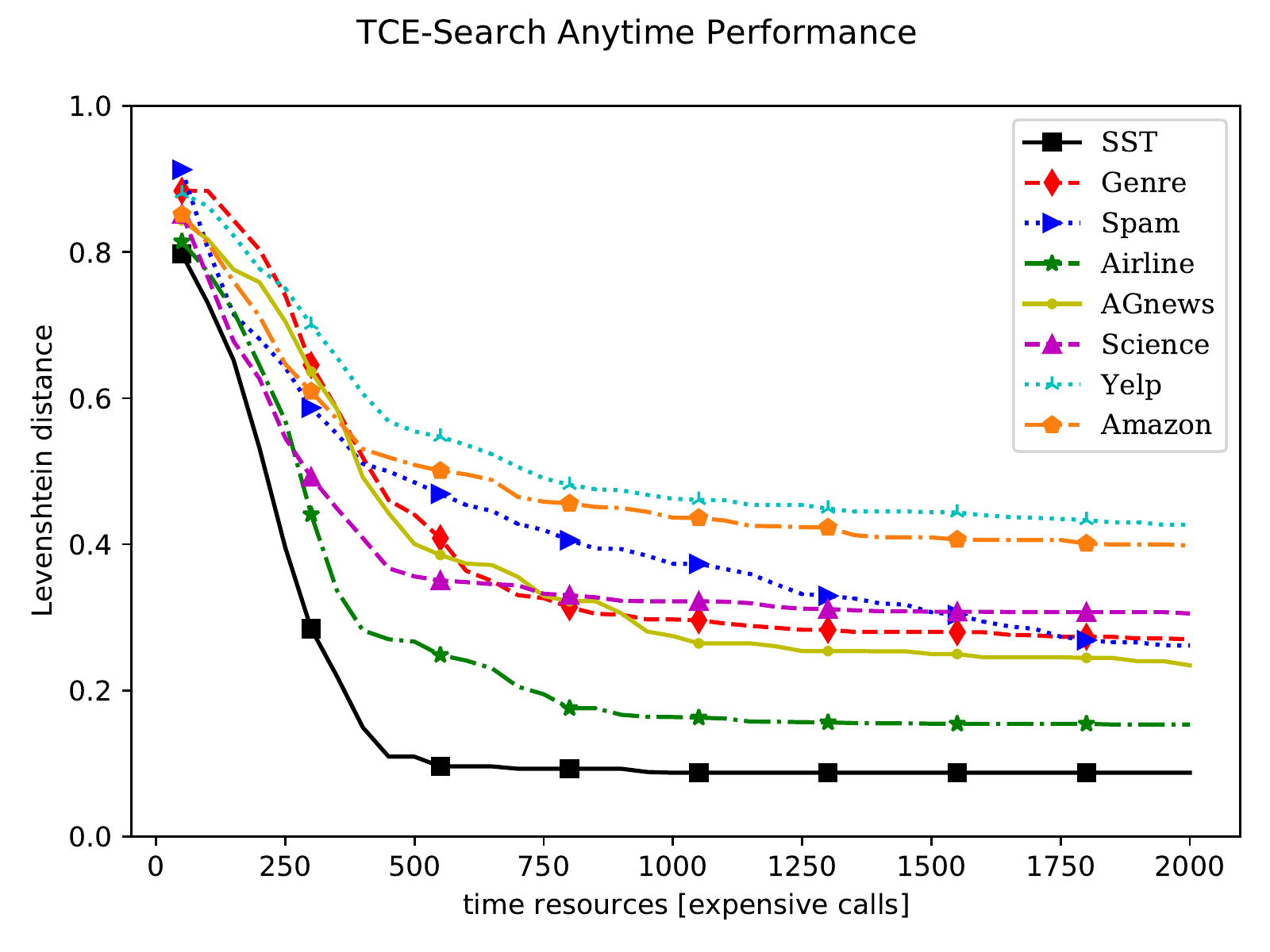}
    \caption{Anytime performance: Levenshtein distance as a function of allocated expensive calls (EC).}
    \label{fig:anytime_performance}
\end{figure}

\subsection{Anytime Performance}

As mentioned in Section \ref{the_search_framework}, \algname\ is an anytime algorithm: It returns a higher quality solution with increase in allocated time. This is important for our use case, where the explanation is likely needed during an interactive session  \cite{nielsen1994usability}.

The anytime performance of \algname\ can be observed in Figure \ref{fig:anytime_performance}. The results demonstrate that for all 8 datasets, as more resources are allocated, the average normalized distance decreases until a stable point is reached. The plots
indicate that \algname\ can generate valid counterfactuals quickly when resources are limited but can  also improve the quality of counterfactuals as more resources become available.

\subsection{Robustness to Different Classifiers} \label{tab:classifier_robustness}
To demonstrate the model-agnostic nature of \algname\, we have tested its performance using 3 different classifiers that used 3 different embedding methods\footnote{The scikit-learn library \cite{scikit-learn} was used for models and embeddings.}: (1) A random forest classifier with CountVectorizer, used in most previous experiments\footnote{When comparing with MiCE (section \ref{section:comparison_with_existing_works}) we used RoBERTa-LARGE.}. (2) A logistic regression classifier with HashingVectorizer. (3) A SVM classifiers with TfidfVectorizer. 
Details on the models training and test set performance can be found in the appendix.

Table \ref{tab:classifier_robustness} illustrates that \algname\ is relatively consistent in its performance across different embedding techniques and classification algorithms, at least within the scope of this experiment.

\begin{table*}[t]
  \centering
  \small
  \caption{Average normalized Levenshtein distance, for \algname\ with various black-box classifiers.}
    \begin{tabular}{lccccccccc}
        \toprule

        Classifier & SST & Yelp & Amazon & Genre & Spam & Airline & AGnews & Science\\
         \toprule
         RF & 0.088 & 0.426 & 0.398 & 0.269 & 0.261 & 0.153 & 0.234 & 0.305
         \\
         LR & 0.102 & 0.525 & 0.386 & 0.431 & 0.293 & 0.214 & 0.296 & 0.413
          \\
          SVM & 0.111 & 0.617 & 0.48 & 0.448 & 0.335 & 0.215 & 0.32 & 0.428
          \\
    \bottomrule
   \end{tabular}

  \label{tab:classifier_robustness}
\end{table*}

\begin{table}
  \centering
  \caption{Language modeling loss ratio between \algname\ generated counterfactuals and original examples over explanation test sets.}
    \resizebox{0.49\textwidth}{!}{\begin{tabular}{cccccccc}
        \toprule
        \multicolumn{8}{c}{Plausibility Ratio}\\
        \cmidrule(rl){1-8}
           Yelp             & SST             & Amazon             & Genre             & Science             & AGNews         & Airline      &Spam   \\
        \cmidrule(rl){1-8}
          1.05         & 1.03          & 1.02        & 1.04           & 1.07        & 1.2      & 1.09     & 1.05   \\
        \bottomrule
    \end{tabular}}
  \label{tab:plausibility_ratio}
\end{table}

\subsection{The Plausibiliy of Generated Counterfactuals}

As mentioned in Section \ref{section:plausibility_enforcement}, in order to ensure plausibility, we monitor the LM loss ratio of candidate states during the search process, and allow only candidates with a loss ratio below the $\gamma$ threshold (1.5 in our experiments).

Similarly to \citet{ross2020explaining} and \citet{ross-etal-2022-tailor}, we first evaluate the plausability of the generated counterfactuals using their measured average LM loss ratio.  The results are presented in Table \ref{tab:plausibility_ratio}. As can be seen, the actual average ratios are significantly lower than $\gamma$ and are close to a ratio of 1.0, which would mean equal LM loss (indicating equal plausibility) for the counterfactual and the original example.

We further evaluated the plausabilty using 
human raters. We randomly sampled 50 examples of the SST explanation test set, and asked  non-native English speakers with a high level of proficiency in the language to rank the plausibility of each of the 50 original examples, and the 50 counterfactuals generated by \algname\ for these texts. We used 3 annotators for each example. The annotators were asked: "Please rank each text on a scale of 1 to 5, where 1 indicates that the text is not a movie review, and 5 indicates that it is a movie review".

We averaged the 3 labels for each text. The average score for the original texts was 3.66, and for the generated counterfactuals 3.33. A paired t-test ($p=0.05$) has shown that the difference is not statistically significant. The average standard deviation between the 3 labels of the original texts is 0.749, and for the counterfactuals generated by \algname\ is 0.643.

\subsection{Multi-Class Setting} \label{section:multi_class}

The experiments in this paper have focused on binary classification problems for simplicity. However, \algname\ is also well-suited multi-class classification tasks. To demonstrate this, we modified the Genre and AGnews datasets to have three classes, as outlined in the appendix, and tested \algname's performance on them. We generated 2 counterfactual examples for each sample, one for every possible target label. The average distance for the Genre explanation test set was 0.57, and 0.257 for the AGnews.

\section{Related Work} \label{section:related_work}

Research from the social sciences suggests that 
often, when humans provide explanations as to why an event had occured, they do not explain the cause to the event per se, but rather explain the cause compared to some other event that did not occur 
\cite{miller2019explanation, Alvarez2019, hilton1990conversational}. The common term to this kind of explanations is \emph{counterfactual explanations}. 

There are various cases where counterfactual explanations are better than feature-importance based explanations \cite{Fernandez20, ross2020explaining}. 
In recent years, therefore, researchers 
have developed algorithms for 
generation 
of counterfactual explanations. One of the first works to propose counterfactual explanations to decisions made by black-box classifiers was by \citet{wachter2017counterfactual}. Several works have proposed to perturbe the original example in the feature space \cite{karimi2020model, mothilal2020explaining, galhotra2021explaining}. 
While these methods are of great use in some domains, they may be inapplicable in other important domains, such as computer vision and NLP, where points in the feature space are incomprehensible to humans \cite{zareckitextual}. 

Recent research has investigated the concept of generating counterfactuals in the textual domain. These methods often utilize transformer-based generative language models \citep{vaswani2017attention}, which have a limitation on the maximum length of input sequences. Some prevalent approaches are designed to modify single sentences. \citet{DBLP:conf/aaai/MadaanPPS21} generate counterfactuals for sentences in order to produce test cases and debias the model. Polyjuice \cite{DBLP:conf/acl/WuRHW20} is a general-purpose counterfactual creator, which generates modified sentences with small syntactic and semantic distance to the original example. Tailor \citep{ross-etal-2022-tailor} modifies sentences in a semantically-controlled fashion, using control codes derived from semantic representation. \citet{robeer2021generating} generates perceptibly distinguishable (large semantic distance compared to original example) counterfactuals, in a model agnostic manner.

Other approaches require a white box access to the model \cite{yang2020generating, DBLP:conf/emnlp/FernP21}. The frameworks suggested by \citet{jacovi2021contrastive} and  \citet{yang2021generative} are limited to neural models.

 MiCE \cite{ross2020explaining} and CAT \cite{DBLP:CAT} are similar to \algname\  in utilizing a MLM to generate plausible counterfactuals. Unlike \algname\ , however,  these methods are not completely model agnostic, as they use the black box's
gradient information to select which parts of the text to mask.  This
limits their use to differentiable models.

\citet{DBLP:conf/dis/LampridisGR20} builds a decision tree in the latent neighborhood of the original instance, and then extracts exemplars and counter-exemplars, close to the original example according to Euclidean distance.
\citet{dixit2022core} generates counterfactuals for the purpose of data augmentation, by constructing detailed prompts to GPT-3 \cite{brown2020language}.

Another limitation of methods that are based on generative language models is that they are optimized for a specific loss function, which may not be suitable for other tasks. Therefore, in order to use these models for other tasks, retraining is often required, which can be costly. In contrast, \algname\ has the ability to adapt to different user needs and constraints without the need for retraining.

We are aware of no other work that generates counterfactual explanations for text classification in an anytime, model-agnostic and domain-agnostic manner, with the ability to handle texts of any length while being able to adapt to varying user requirements without retraining.

\citet{MartensP14}, \citet{Fernandez20} and \citet{Ramon19} generate explanations by selecting the words in the text that are
important for the classification.  While they use the term  "counterfactual explanations", their definition of such explanations is different from the common usage that refers to a complete object of the opposite class.

Note that while \emph{adversarial learning}
algorithms also perturb examples, their goals and constraints are completely different \citep{verma2020counterfactual}. The aim in adversarial learning is to fool the classifier, i.e,
generating an example whose 
true label remains the same as of the original example, but the label given to it by the classifier is different.  The problem of counterfactual explanation generation is not subject to this constraint.
On the other hand, adversarial learning algorithms are not restricted 
to generating plausible and human understandable objects,  hence they cannot be used for explanation.

\section{Conclusion} \label{section:conclusion}

In this paper, we presented \algname, a general model-agnostic search-based framework for generating counterfactual explanations for text classification. Our approach adapts to user-specified constraints and preferences, such as a distance function and computational budget, \emph{without retraining}. \algname's counterfactuals are plausible, valid and close to the original examples, outperforming current state-of-the-art approaches.

Future work should investigate the \emph{utility} of these explanations to humans through further user studies and controlled experiments.

\bibliography{main}

\clearpage
\appendix

\section{Data Pre-Processing}
We used binary datasets for most experiments in this paper, for simplicity. In the pre-processing stage, we made the necessary adjustments to transform the datasets to binary-class datasets:

For the Yelp\footnote{https://www.yelp.com/dataset} and Amazon datasets, we labeled reviews with stars $\geq$ 4 as \emph{positive}, reviews with stars $\leq$ 2 as \emph{negative}, and dismmissed the other reviews (the "neutrals"). Similarly, for the SST dataset we labeled examples with rating $\geq$ 0.6 as \emph{positive}, those with rating $\leq$ 0.4 as \emph{negative}, and discarded the other examples.
We also removed tweets labeled as "neutral" from the Airline\footnote{https://www.kaggle.com/crowdflower/twitter-airline-sentiment} dataset, the remaining are already classified as \emph{positive} or \emph{negative}.

For the Science dataset\footnote{https://www.kaggle.com/vivmankar/physics-vs-chemistry-vs-biology}, we removed texts about chemistry, and remained with a binary dataset of either physics or biology. In a similar manner, we removed texts from the Genre dataset\footnote{https://www.kaggle.com/hijest/genre-classification-dataset-imdb?select=Genre+Classification+Dataset} whose label is not "drama" or "comedy", and reduced the AGNews dataset to texts labeled either "business" or "sports".  The Spam dataset is binary.

For the experiment described in Section 4.7, we also allowed texts with "documentary" labels for Genre, and texts with "tech" labels for AGnews, to create a multi-class setting.

In a class-unbalanced setting, a weakness of \algname\ in one direction of transformation could have been camouflaged. To avoid such a biased evaluation, we balanced the datasets by randomly undersampling the majority class.

\section{Black-Box Training and Performance}
For most of the experiments described in this paper, 
we used a random forest model with 1000 estimators as the black box, and 
sklearn's CountVectorizer, with max features set to 1500, min document frequency to 5 (total word count) and max document frequency to 0.7 (proportion of corpus), for embedding. 

For the experiments described in Section 4.5, we trained two additional classifiers: a logistic regression classifier that used HashingVectorizer and a SVM classifier that used TfidfVectorizer. For these  classifiers and embedding methods, the default hyperparameters provided by scikit-learn were used.

The models' accuracy scores across the eight black-box test sets are available on Table \ref{tab:classifiers_acc}.

\section{Text Cleaning}
We cleaned both the original example and the generated counterfactual before measuring the distance. The cleaning procedure included removing redundant spaces before punctuation (e.g. " ," was replaced by ",") and removing the chars '\#' and '@').

\begin{table}
  \centering
  \caption{Test sets accuracy for various black-box classifiers.}
    \resizebox{0.49\textwidth}{!}{\begin{tabular}{lccccccccc}
        \toprule

        Classifier & SST & Yelp & Amazon & Genre & Spam & Airline & AGnews & Science\\
         \toprule
         RF & 0.733 & 0.905 & 0.864 & 0.751 & 0.959 & 0.843 & 0.947 & 0.779
         \\
         LR &  0.779 & 0.917 & 0.862 & 0.776 & 0.955 & 0.871 & 0.961 & 0.785
          \\
          SVM & 0.804 & 0.935 & 0.878 & 0.799 & 0.96 & 0.891 & 0.973 & 0.829
          \\
    \bottomrule
   \end{tabular}}
   \label{tab:classifiers_acc}
\end{table}

\begin{table}
  \centering
    \caption{Examples of \algname\ counterfactuals.}
    \begin{tabular}{p{0.2\textwidth}p{0.2\textwidth}}
        \toprule
          Original Text                & Counterfactual      \\
        \midrule
          illinois apparently is going to spend a little longer at no. 1 this time.  (Sports)       & \textbf{company} apparently is going to spend a little longer at no. 1 this time. (Business)      \\ \midrule
          ohio holdings corp. (Business) &   ohio holdings \textbf{team}. (Sports) 
              \\ \midrule
           the latest adam sandler assault and possibly the worst film of the year. (Negative) &
           the latest adam sandler assault and possibly the \textbf{best} film of the year. (Positive) \\ \midrule
            like halo? then you'll like halo 4. (Positive) & 
            like halo? then you'll \textbf{not} like halo 4. (Negative)\\ \midrule
            what a great system!. the 4gb is not very much if you want to download anything. use a flash drive or buy a dedicated hard drive. (Positive) & 
            what a \textbf{crap} system!. the 4gb is not very much if you want to download anything. use a flash drive or buy a dedicated hard drive. (Negative) \\
        \bottomrule
    \end{tabular}
  \label{tab:more-example-cfs}
\end{table}

\end{document}